\pdfoutput=1
\documentclass[]{style}

\titlecardtrue

\title{Hard Vision, Easy Vision: What GPT-6 Astra Reveals Across Computer Vision}

\toplogo[trim=12.23bp 12.38bp 18.98bp 12.23bp, clip]{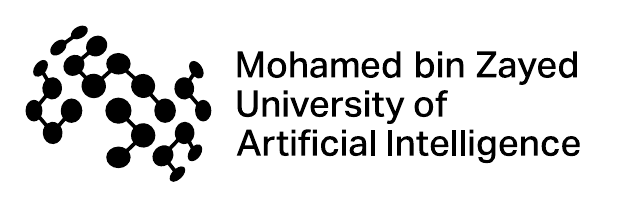}

\author[1,\textdagger]{Hanoona Rasheed}
\author[1,\textdagger]{Mohammed Irfan Kurpath}
\author[1]{Bin Ren}
\author[1]{Hisham Cholakkal}
\author[1,2]{Fahad Shahbaz Khan}
\author[1,2]{Salman Khan}

\affiliation[1]{Mohamed bin Zayed University of Artificial Intelligence}
\affiliation[2]{Apertix}
\affiliation[\textdagger]{Equal contribution}

\date{Sep 28, 2026}
\metadata[Project Page]{\href{https://mbzuai-oryx.github.io/frontier-vision}{\nolinkurl{mbzuai-oryx.github.io/frontier-vision}}}
\metadata[GitHub]{\href{https://github.com/mbzuai-oryx/frontier-vision}{\nolinkurl{mbzuai-oryx/frontier-vision}}}

\usepackage[dvipsnames,table]{xcolor}
\usepackage{amsmath}
\usepackage{amssymb}
\usepackage{array}
\usepackage{longtable}
\usepackage{float}
\usepackage{pdflscape}
\usepackage{enumitem}
\usepackage{xspace}
\usepackage{tikz}
\usepackage{fontawesome5}

\newsavebox\fittobox
\newcommand{\fitto}[2]{\sbox\fittobox{#2}\ifdim\wd\fittobox>#1\resizebox{#1}{!}{\usebox\fittobox}\else\usebox\fittobox\fi}
\definecolor{tabF4867C}{HTML}{F4867C}
\definecolor{tabF7AE85}{HTML}{F7AE85}
\definecolor{tabF9D38B}{HTML}{F9D38B}
\definecolor{tabE6EE8E}{HTML}{E6EE8E}
\definecolor{tabBDE69A}{HTML}{BDE69A}
\definecolor{tab86DCA1}{HTML}{86DCA1}
\definecolor{tab53C98C}{HTML}{53C98C}
\definecolor{grp1F8AFF}{HTML}{1F8AFF}
\definecolor{grp986BF6}{HTML}{986BF6}
\definecolor{grpD152BB}{HTML}{D152BB}
\definecolor{grpEC4764}{HTML}{EC4764}
\definecolor{grpFF8506}{HTML}{FF8506}
\definecolor{grpF2CB04}{HTML}{F2CB04}
\definecolor{grp59AA03}{HTML}{59AA03}
\definecolor{grp00B89A}{HTML}{00B89A}
\definecolor{grp06B1CE}{HTML}{06B1CE}

\newcommand{\vs}{vs.\xspace}

\newlist{checklist}{itemize}{1}
\setlist[checklist]{label=$\square$}

\newcolumntype{L}[1]{>{\raggedright\arraybackslash}p{#1}}

\usepackage{tikz}
\usepackage{xcolor}
\definecolor{rqblue}{HTML}{0055BE}
\definecolor{rqbg}{HTML}{E8F5FD}
\usetikzlibrary{calc,backgrounds}
\newcommand{\researchquestion}[3][]{%
\par\vspace{-6pt}%
\noindent
\begin{tikzpicture}
    \node[
        anchor=north west,
        inner xsep=0pt,
        inner ysep=3.2mm,
        minimum height=12mm,
        text width=\dimexpr\linewidth-12mm\relax
    ] (rq) at (5mm,0) {%
        \textcolor{rqblue}{\textbf{#2:}}\, #3%
        \if\relax\detokenize{#1}\relax\else
          \par\vspace{3pt}{\small\textcolor{rqblue}{\textbf{TL;DR:}}\, #1}%
        \fi
    };
    \begin{scope}[on background layer]
      \path let \p1 = (rq.south) in
          [fill=rqbg]
          (0,0)
          -- (\linewidth-3mm,0)
          arc[start angle=90,end angle=0,radius=3mm]
          -- (\linewidth,\y1+3mm)
          arc[start angle=0,end angle=-90,radius=3mm]
          -- (0,\y1)
          -- cycle;
      \path let \p1 = (rq.south) in
          [draw=rqblue, line width=2.5pt] (0,0) -- (0,\y1);
    \end{scope}
\end{tikzpicture}
\par\vspace{-6pt}%
}

\abstract{
Frontier general-purpose systems are rapidly expanding beyond visual understanding into capabilities traditionally handled by dedicated computer-vision models. 
As these capabilities expand, a central question for the computer-vision community is how far this reach extends, and what remains hard.
We evaluate GPT-6 Astra alongside five frontier general-purpose AI systems across 34 capabilities and 55 benchmarks spanning nine areas of computer vision.
We compare their performance with dedicated models and humans where suitable references are available.
Astra demonstrates broad visual capability, with substantial gains over other frontier systems in visual and spatial reasoning and several forms of structured prediction. 
Across the state-of-the-art systems, a consistent pattern emerges. Capabilities involving semantic interpretation, reasoning, and object-centric prediction increasingly approach or reach available reference levels.
In contrast, larger gaps remain when tasks require metric geometric accuracy, faithful reconstruction, temporally consistent dense prediction, or specialized fine-grained visual knowledge. 
Additional reasoning and specialist tools close selected gaps, but their benefits vary across capabilities. 
These results map a changing landscape of computer vision in which increasingly sophisticated visual tasks are accessible through a general-purpose interface, while precise and fidelity-sensitive perception remains an important frontier.
}

\begin{document}

\maketitle

\begin{figure}[H]
\vspace{-3pt}
  \centering
  \includegraphics[width=\linewidth]{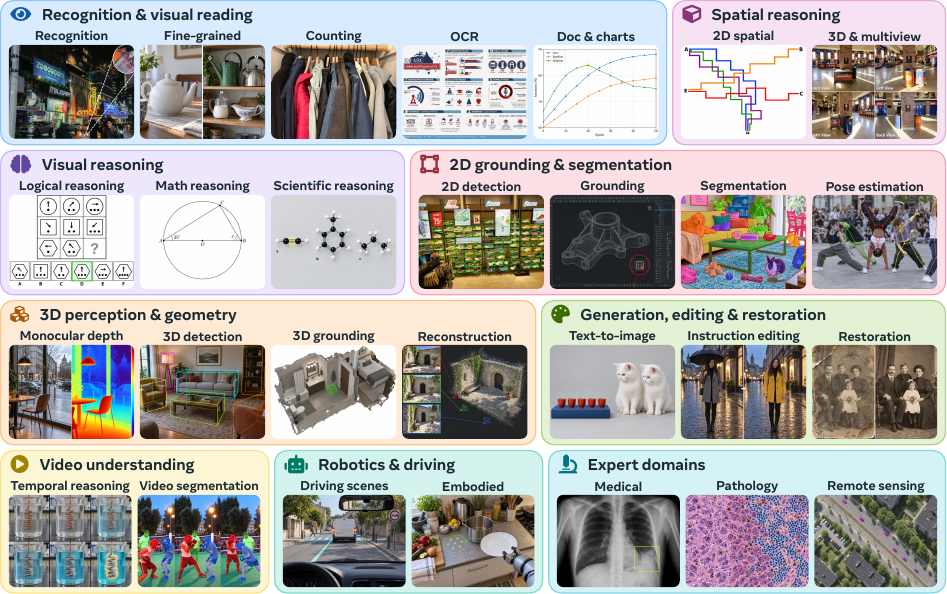}
  \vspace{-13pt}
  \caption{\textbf{Mapping the changing landscape of general-purpose vision.} Our study covers 34 capabilities across nine broad areas of computer vision, drawing on 55 benchmarks; representative tasks are illustrated here. We compare frontier systems with specialist models and human performance to examine how much of computer vision is now accessible through a general-purpose interface, where meaningful gaps remain, and where dedicated vision models are still necessary.}
  \label{fig:hero}
\end{figure}

\section{Introduction}
\label{sec:intro}

Computer vision has traditionally advanced through specialized models trained for individual tasks: classifiers for recognition~\citep{krizhevsky2012imagenet,oquab2023dinov2}, detectors~\citep{ren2016faster,liu2024grounding} and segmentation models~\citep{long2015fully,kirillov2023segment} for localization, geometric models for depth~\citep{eigen2014depth, yang2024depth} and 3D perception~\citep{qi2017pointnet,wang2025vggt}, video models~\citep{simonyan2014two,assran2025v} for temporal understanding, and domain-specific models for areas such as robotics and medical imaging~\citep{ronneberger2015u,ma2024segment}. \textit{This landscape is now changing.} Frontier general-purpose systems increasingly bring diverse visual capabilities into a shared interface allowing users to specify tasks through natural-language instructions~\citep{liu2023visual,openai2023gpt4v,team2023gemini}. Their expanding reach beyond semantic image understanding into structured prediction, visual generation, and embodied interaction is changing expectations of what general-purpose models can accomplish. As new capabilities emerge, a fundamental question becomes increasingly relevant to the field: \textit{\textbf{how much of computer vision is now accessible through a general-purpose system, and where do dedicated vision models remain necessary?}}

\begin{table}[!t]
  \centering
  \resizebox{\linewidth}{!}{\input{sections/tables/capability_table_gradient}}
  \caption{\textbf{Frontier visual capability: shared strengths, differences, and remaining headroom.} We compare six frontier general-purpose systems across \textbf{34 capabilities spanning nine broad areas of computer vision}. \textbf{Reference Model} reports either a \textit{dedicated specialist}, trained or designed specifically for the task, or a \textit{prior generalist SOTA}, an earlier general-purpose model with the strongest reported result on the benchmark. \textbf{Human} is based on benchmark-reported human performance where available. The last two columns summarize complementary aspects of the capability landscape: Best model vs. ref. measures how far the best generalist exceeds or trails the strongest available model or human reference, indicating the remaining headroom (RQ3); Best vs. 2nd best reports its margin over the next-best generalist, highlighting where a particular system stands out (RQ2). Positive gaps indicate better performance, with signs adjusted for $\downarrow$ metrics.}
  \label{tab:capabilities-abs}
  \vspace{-10pt}
\end{table}

Existing benchmarks~\cite{yue2025mmmu,fu2024blink} provide extensive evaluations of visual tasks and a substantial body of evidence about the capabilities of frontier models. This evidence, however, is spread across tasks, domains, and model comparisons, making it difficult to see what individual advances collectively mean for computer vision. A model may lead other generalists on a benchmark while remaining far from specialist or human performance~\citep{tong2024eyes,fu2024blink}, and its strengths on one task may not extend to related tasks. Connecting and interpreting these results through a landscape-level analysis can reveal which capabilities are becoming broadly accessible, how close they are to established reference levels, and where meaningful gaps remain. Understanding this landscape can clarify the evolving role of specialized models, identify capabilities with substantial remaining headroom, and guide future computer-vision research toward the challenges where it can make the greatest difference.

To investigate this question, we conduct a systematic study of the breadth and limits of frontier general-purpose vision. We compare systems with one another and examine how their performance relates to that of dedicated vision models and humans. We organize our analysis around four research questions: \textbf{(RQ1)} How broad is the visual coverage of current frontier systems relative to human and specialist references? \textbf{(RQ2)} Where do frontier systems converge, and where do they still differ substantially? \textbf{(RQ3)} For which task types is specialist-level performance available through a general-purpose interface, and what task properties predict the remaining gap? \textbf{(RQ4)} Can added reasoning, explicit tool use, or open specialist-as-tool pipelines close those gaps, and at what cost? Across these evaluations, we observe a consistent boundary emerging: general-purpose systems like GPT6-Astra increasingly match reference performance when visual information supports semantic interpretation and reasoning. In contrast, the largest gaps persist when the output must remain metrically precise, pixel-faithful, temporally consistent, or dependent on specialized fine-grained visual knowledge.

\section{Mapping the Computer Vision Landscape and Evaluation Setup}
\label{sec:setup}

We organize the evaluation into \textbf{34 capabilities} spanning \textbf{nine broad areas} of computer vision: \textit{i}{\scriptsize)} recognition, perception, and visual reading; \textit{ii}{\scriptsize)} visual reasoning; \textit{iii}{\scriptsize)} spatial reasoning; \textit{iv}{\scriptsize)} 2D grounding, detection, and segmentation; \textit{v}{\scriptsize)} 3D perception and geometric prediction; \textit{vi}{\scriptsize)} video understanding and segmentation; \textit{vii}{\scriptsize)} image generation, editing, and restoration; \textit{viii}{\scriptsize)} robotics; and \textit{ix}{\scriptsize)} expert-domain vision.
Across these capabilities, we draw on \textbf{55 benchmarks}, prioritizing challenging evaluations that retain meaningful headroom for current frontier systems while collectively covering a broad range of computer-vision tasks. Our goal is to evaluate not only what these systems can understand from visual inputs, but also the range and precision of the outputs they can produce. The resulting tasks therefore extend beyond textual answers to bounding boxes and masks, depth maps and 3D predictions, temporally consistent mask sequences, generated and edited images, and actions in embodied environments. Together, these evaluations capture \textbf{a broad range of visual capabilities}, from semantic understanding to precise structured prediction and task execution.

We evaluate \textbf{six frontier general-purpose systems}: GPT-6 Astra~\citep{openai2026gpt6astra}, Fable 5~\citep{anthropic2026fable5}, Kimi K3~\citep{team2026kimi}, Gemini 3.1 Pro~\citep{google2026gemini31pro}, Qwen 3.8-Max~\citep{qwen38}, and Muse Spark 1.3~\citep{meta2026musespark}. All models receive the same task instructions, visual inputs, and evaluation samples on each benchmark, with outputs scored using the corresponding benchmark metric. Where models provide multiple reasoning configurations, we use the strongest available reasoning setting appropriate to the task. This establishes a common evaluation basis for examining both capabilities increasingly shared across frontier systems and those where substantial differences remain.

To assess how far general-purpose capability has progressed, we compare frontier systems against both \textit{specialist} and \textit{human} performance where suitable references are available. Specialist references are drawn from leading task-specific methods whose architectures, training procedures, or optimization are designed specifically for the corresponding capability, providing a measure of how closely general-purpose systems approach performance achieved by dedicated vision models. Human performance provides a complementary reference for understanding the remaining headroom beyond specialist-level capability. To characterize how far each capability has progressed toward these reference levels, we describe performance using four maturity tiers: \textit{exceeds reference level}, \textit{at reference level}, \textit{approaching}, and \textit{substantial gap}. These tiers summarize capability maturity on the evaluated benchmarks rather than implying that the underlying task itself is solved.

\begin{figure}[t]
  \centering
  \includegraphics[width=\linewidth]{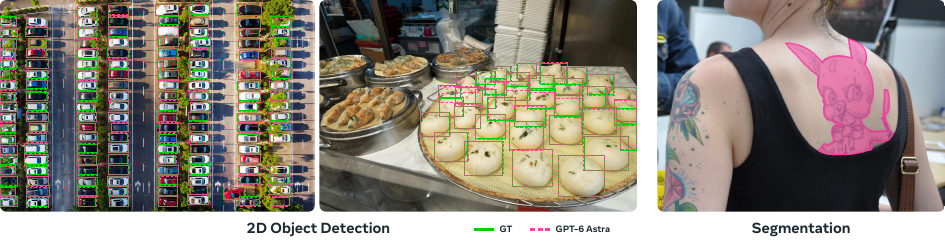}
  \vspace{-15pt}
  \caption{\textbf{Specialist-level 2D localization through a general-purpose interface.} We illustrate this capability with GPT-6 Astra, which detects densely packed objects in crowded scenes (left) and captures the intricate contours of a tattoo through segmentation (right), showing structured prediction capabilities traditionally handled by dedicated vision models.}
  \label{fig:det_seg}
\end{figure}
\begin{figure}[!t]
  \vspace{-6pt}%
  \centering
  \includegraphics[width=\linewidth]{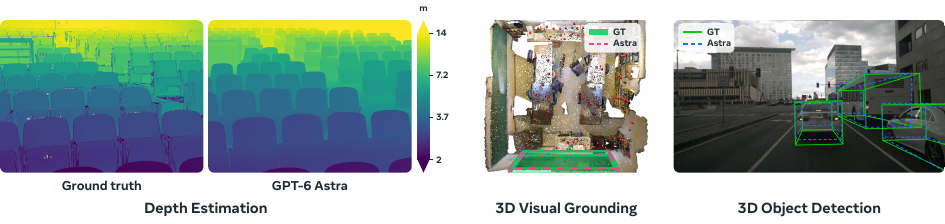}
  \vspace{-15pt}
  \caption{\textbf{Advances in object-centric 3D understanding and remaining geometric gaps.} GPT-6 Astra accurately localizes the referred blackboard (middle), illustrating a capability where general-purpose performance now reaches and even exceeds specialist levels, while showing a large margin over other frontier models (+10.7). It also extends to 3D object detection, with coherent bounding boxes (right). Depth prediction, however, remains more challenging: although the depth map captures the broad scene layout, it smooths over fine surface and boundary details (left), highlighting the remaining headroom in precise geometric prediction.}
  \label{fig:astra_3d_tasks}
\end{figure}

\section{How Broad Is Frontier Visual Capability?}
\researchquestion[Across 34 capabilities and 55 benchmarks covered in this study, we notice that the frontier has broadened dramatically. However, breadth is not uniform in maturity.]{Question 1}{\textbf{How broad is the visual coverage of current frontier systems relative to human and specialist reference performances?}
}
\textbf{Recognition, Perception, and Visual Reading}.\, The perception results show two notable trends. \textit{i}{\scriptsize)} Strong perception is increasingly a shared property of frontier models. All six evaluated models perform well on challenging tasks spanning visual recognition, fine-grained discrimination and matching, object counting, and visual reading. \textit{ii}{\scriptsize)} Visual reading approaches or exceeds human performance. Compared with recognition and counting, OCR comes closer to human-level performance (93.7--98.1 vs.\ 98.0). Document, chart, and infographic understanding is even stronger: all six models surpass the reported human performance, by margins ranging from +0.2 to +3.5 points.

\textbf{Visual and Spatial Reasoning}.\, In contrast to perception, visual and spatial reasoning remain less uniformly mature across frontier models. Within visual reasoning, predominantly visual logical problems retain more headroom, while tasks that combine visual information with mathematical and scientific knowledge are closer to human performance. All six models reach or exceed the human score on mathematical reasoning (78.8--92.5 vs.\ 78.7), while scientific and professional reasoning remains slightly below it (80.5--86.8 vs.\ 88.6). Another key observation is the \textit{emergence of spatial reasoning as a strong capability} in the latest frontier models, with GPT-6 Astra reaching the human performance in 2D spatial reasoning (96.0 vs.\ 95.8) and approaching it in 3D spatial and multiview reasoning (89.6 vs.\ 94.1).

\begin{figure}[t]

  \centering
  \includegraphics[width=\linewidth]{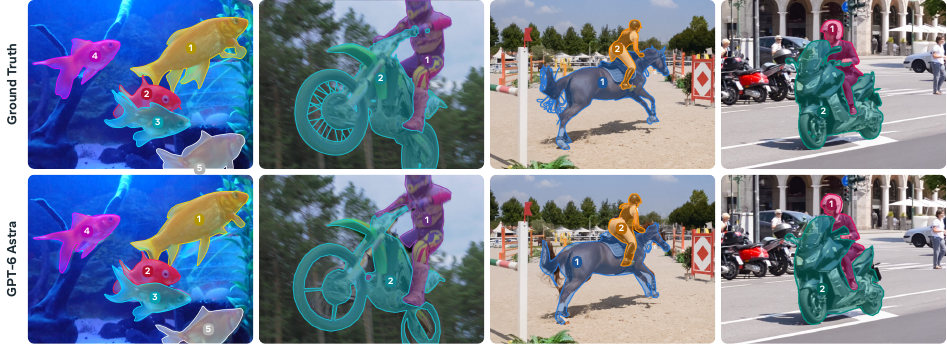}
  \vspace{-15pt}
  \caption{\textbf{Remaining gaps in dense temporal prediction}. To illustrate the remaining challenges in video segmentation, we show outputs from GPT-6 Astra, which achieves substantial gains over the next-best frontier system (+26.6 points). The predictions capture the main object shapes and distinguish multiple instances, while precise boundaries, small structures, and separation of nearby instances remain challenging. These examples highlight the spatial distinctions that must be maintained consistently across frames to close the remaining specialist gap.}
  \label{fig:davis_qualitative}
\end{figure}

\begin{figure}[t]
  \centering
  \includegraphics[width=\linewidth]{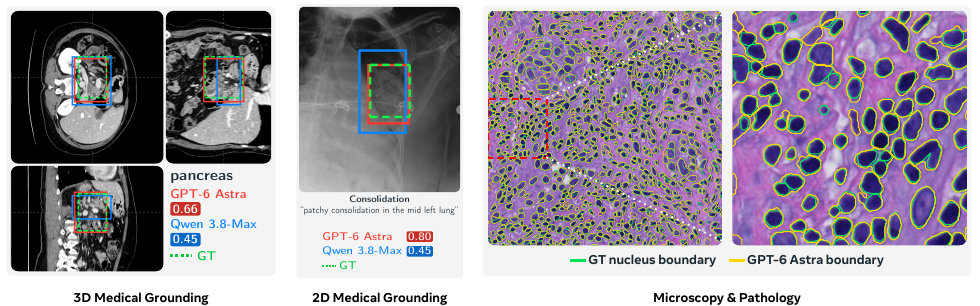}
\caption{\textbf{Emerging medical grounding capabilities and remaining gaps in domain expertise.} In 3D medical grounding, GPT-6 Astra accurately localizes the pancreas in abdominal CT, with predictions shown in axial (top left), sagittal (right), and coronal (bottom) views. In 2D medical grounding, successful localization is also possible, although substantial headroom remains relative to dedicated models. The pathology example, shown in full view and zoom-in, demonstrates accurate nucleus localization and delineation. Yet distinguishing fine-grained nucleus types remains substantially harder, highlighting the domain expertise required to interpret these structures.}
  \label{fig:medical}

\end{figure}

\textbf{Localization, 3D Perception, and Video Understanding}.\, \textbf{\textit{i)} 2D}: The results show localization emerging as a strong capability in the latest frontier models, extending into tasks traditionally handled by specialized models. GPT-6 Astra exceeds the specialist performance in both object detection and segmentation (+10.7 and +4.3 points), while its visual grounding performance approaches the human performance (93.1 vs.\ 96.0) (See \Cref{fig:det_seg}). This trend extends beyond a single model, with Qwen 3.8-Max also surpassing the specialist detector (+8.6 points), indicating a broader shift toward specialist-level 2D localization through general-purpose models. \textbf{\textit{ii)} 3D}: Progress is less uniform in 3D perception. Object-related capabilities show stronger progress toward specialist performance: all six models outperform the specialist in 3D visual grounding, while on the distinct task of 3D object detection, GPT-6 Astra closely approaches the specialist performance (17.3 vs.\ 16.8) (See \Cref{fig:astra_3d_tasks}). In contrast, metric depth and reconstruction retain substantial headroom; even the strongest multiview reconstruction result has approximately $2.3\times$ the error of the specialist model. \textbf{\textit{iii)} Video}: A similar distinction appears between video understanding and dense temporal prediction. Several frontier models are already competitive with the specialist in video and temporal understanding (72.6--76.3 vs.\ 73.3). However, this competitiveness does not yet extend to video segmentation, where even the best-performing frontier model remains 7.9 points below the dedicated model (See \Cref{fig:davis_qualitative}). Overall, these results point to 2D localization and object-centric 3D understanding as \textit{emerging strengths} of general-purpose models, while precise geometric prediction and temporally consistent dense prediction continue to show substantial headroom.

\begin{figure}[t]
  \centering
  \includegraphics[width=\linewidth]{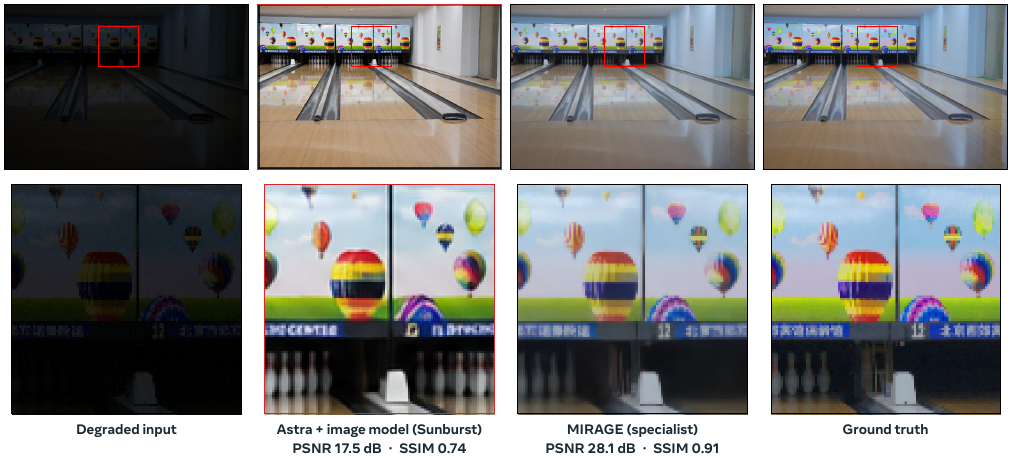}
  \caption{\textbf{Astra's restorations look plausible but hallucinate content}. LOL-v1 low-light enhancement (t2p-001789). Columns: degraded input, Astra + image model (Sunburst), MIRAGE (specialist), ground truth; the bottom row enlarges the red boxes. Left: four bowling pins become five, extra pins appear on an empty lane, and the banner text and lane number are altered, while MIRAGE stays faithful. Such hallucinations may be acceptable for casual photos but would be critical in fidelity-sensitive domains such as medical imaging. PSNR/SSIM are computed on full images.}
  \label{fig:image-restoration}
\end{figure}

\textbf{Image Generation, Editing, and Restoration}.\, \textbf{i)} In generation, editing, and quality assessment, the results show broadly consistent performance across most frontier models, with competitive results close to specialist levels in image generation, instruction-guided editing, and quality assessment. GPT-6 Astra exceeds specialist performance in all three tasks (+3.6, +0.16, and +1.3 points, respectively). \textbf{ii)} The performance observed in generation and editing does not extend to restoration. Here, all evaluated generalist models perform well below the specialist, with scores concentrated in a narrow range (17.2–17.7 vs. 30.7 dB PSNR). This consistency across models indicates that accurate image reconstruction remains a shared limitation, despite their strong generation and editing capabilities. \Cref{fig:image-restoration} illustrates this gap with a low-light enhancement example.

\begin{figure}[t]
  \centering
  \includegraphics[width=\linewidth]{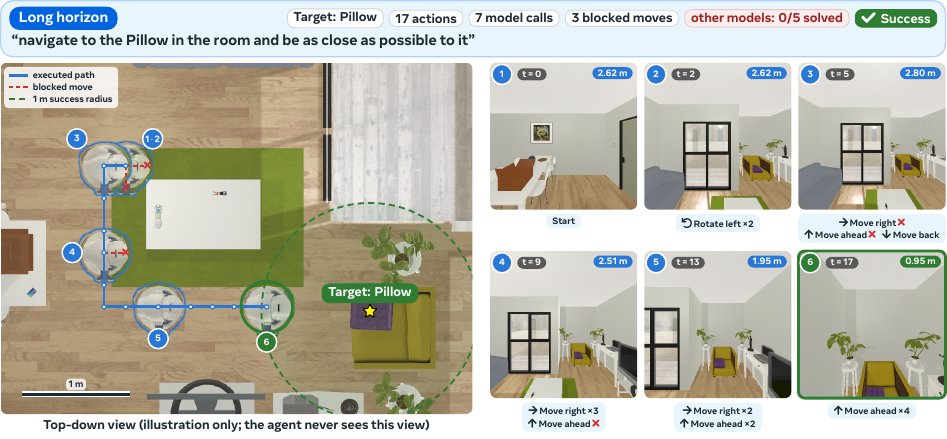}
  \caption{\textbf{Embodied navigation with GPT-6 Astra.} An episode from EB-Navigation (AI2-THOR). Right: egocentric RGB observations at selected steps $t$, the agent's only visual input; it receives no map, target coordinates or privileged simulator state. Chips below each frame list the discrete actions executed since the previous frame, and the badge in each frame gives the remaining distance to the target. Left: a top-down view of the executed path for illustration only, with the 1\,m success radius; the agent never sees this view. Blocked by obstacles, the agent backs up, detours around the obstruction and then approaches the pillow. None of the other five evaluated models solved this episode.}
  \label{fig:embodied_nav}
\end{figure}

\textbf{Robotics and Expert-Domain Vision}.\, \textbf{\textit{i)} Robotics}: The results show competitive driving-scene and embodied understanding across several frontier models. GPT-6 Astra exceeds specialist performance in both tasks (+7.7 and +3.8 points, respectively), while Qwen 3.8-Max and Muse Spark 1.3 also approach specialist-level embodied understanding. Navigation extends this coverage from understanding to the more demanding setting of task execution, with GPT-6 Astra achieving 78\% success in the evaluated setting (\Cref{fig:embodied_nav}). \textbf{\textit{ii)} Expert-domain vision}: Generalist understanding also extends to specialized imagery, including chest radiographs and satellite imagery, with several models approaching or exceeding specialist performance in medical and remote-sensing understanding. However, grounding fine-scale objects in remote-sensing imagery remains challenging, with all evaluated models substantially below specialist performance. Microscopy and pathology reveal a further limitation in domain-specific recognition. Despite their competitive medical-image understanding, all six models remain substantially below both specialist and human performance on these tasks (14.1–23.8 vs. 57.1 and 82.0, respectively). Qualitative observations suggest that models can localize relevant structures yet struggle to distinguish their fine-grained categories, indicating that successful localization does not necessarily imply the domain expertise required to interpret these images.

\textbf{Main takeaway.}\, These results suggest an ongoing competition between \textit{semantic competence} and \textit{perceptual precision}, with frontier systems showing stronger progress in understanding visual content than in measuring or reconstructing it faithfully. Strong 3D grounding coexists with weaker depth estimation and reconstruction; competitive image generation, editing, and quality assessment do not extend to faithful restoration; and strong video understanding does not yet translate into equally strong video segmentation. Similarly, medical understanding is strong, but fine-grained pathology recognition remains weaker.

\section{Where Do Frontier Systems Converge and Differ?}
\researchquestion[Generalists are strongest at semantic interpretation and reasoning; gaps concentrate in metric precision, fidelity, temporal consistency, and specialized fine-grained discrimination.]{Question 2}{\textbf{Where do frontier systems converge, and where do they still differ substantially?}
}
\textbf{Capabilities Where Frontier Systems Converge.}\, \textbf{\textit{i})} The clearest convergence at a strong performance level appears in visual reading, where frontier models achieve consistently high results in OCR (93.7--98.1) and document, chart, and infographic understanding (89.5--92.8). Similar convergence is also visible across conventional perception and reasoning capabilities, including recognition and scientific and professional reasoning, as well as in image generation, instruction-guided editing, image quality assessment, and embodied understanding. The consistently strong performance across multiple systems suggests that these capabilities are increasingly becoming shared strengths of frontier general-purpose models. \textbf{\textit{ii})} Convergence also occurs around shared limitations. Image restoration produces uniformly low scores across these models (17.2--17.7 dB vs.\ 30.7 for the specialist), indicating substantial common headroom. Microscopy and pathology show a similar pattern, with frontier models consistently performing well below the reference (14.1--23.8 vs.\ 82.0). These results distinguish capabilities that are becoming broadly established across frontier general-purpose systems from those where the systems remain uniformly limited.

\textbf{Capabilities Where Substantial Differences Remain.}\, Not all emerging capabilities are yet shared across frontier systems; in several areas, strong performance is concentrated in only a subset of models while others remain substantially behind. In 2D object detection, GPT-6 Astra and Qwen 3.8-Max already exceed the specialist reference (87.0--89.1 vs.\ 78.4), while the remaining models score considerably lower (31.1--69.7). Substantial variation also persists across 3D perception, including pose and depth estimation, 3D detection and grounding, and multiview reconstruction, with different frontier models showing emerging strengths on different subtasks rather than a consistent trend. 
Similar differences remain in expert-domain localization, such as 2D medical grounding. More broadly, \textit{domain reasoning does not necessarily imply domain perception}. A model may have substantial medical knowledge and reason well about medical images, yet lack the perceptual expertise needed to distinguish subtle differences in morphology. The pathology examples illustrate this gap: successful nucleus localization can coexist with difficulty identifying fine-grained nucleus types (See \Cref{fig:medical}).
Video understanding shows another clear emerging cluster: GPT-6 Astra, Qwen 3.8-Max, and Muse Spark 1.3 are already competitive with the specialist reference, while performance across other frontier models still spans a much wider range (53.0--74.3 vs.\ 71.5). Video segmentation shows a particularly large difference across frontier models: GPT-6 Astra achieves 84.5 J\&F, approaching the specialist performance, compared with 32.0--57.9 for the other generalists. Together, these results characterize capabilities that are beginning to emerge strongly in selected frontier systems but have not yet become shared strengths across the frontier.

\textbf{Where New Capability Gains Emerge.}\, Beyond the capabilities that are increasingly shared across frontier models, the strongest signs of further capability expansion appear in visual reasoning and structured prediction. \textbf{\textit{i) Visual Reasoning:}} GPT-6 Astra shows substantial improvements across fine-grained discrimination and matching, visual logical and mathematical reasoning, and 3D spatial and multiview reasoning. Logical and multiview reasoning show two of the largest margins over the next-best frontier models (+13.6 and +12.4 points, respectively). These results highlight reasoning over fine-grained visual information and spatial relationships as emerging strengths beyond the conventional perception capabilities increasingly shared across frontier models. \textbf{\textit{ii) Structured Prediction:}} While several systems already show competitive detection performance, Astra shows substantially larger gains in video segmentation (+26.6 points over the next-best; \Cref{fig:davis_qualitative}), pose estimation (+22.7), image segmentation (+10.4; \Cref{fig:det_seg}), and 3D visual grounding (+10.7; \Cref{fig:astra_3d_tasks}), extending its advantage from image-level structured prediction to dense prediction across video frames.

\section{Where Are Specialist-Level Capabilities Emerging?}
\researchquestion[Specialist-level performance is emerging in semantic and object-centric tasks, while gaps persist in tasks requiring geometric precision, temporal consistency, fine-grained reconstruction, or domain expertise.]{Question 3}{\textbf{For which task types is specialist-level performance available through a general-purpose interface? What task properties predict the remaining gap?}
}
\textbf{Specialist-Level Capabilities Through a General-Purpose Interface.}\, Several capabilities traditionally handled by dedicated vision models are now available through frontier general-purpose models at performance levels comparable to their specialist counterparts. The clearest examples appear in 2D structured prediction, where object detection and segmentation are competitive with specialist models, with similar capability emerging in 3D visual grounding. Beyond localization, comparable performance is also seen in video understanding and embodied settings, including driving-scene and robotic understanding. Image quality assessment shows the same trend. Notably, this reach extends even into expert domains, with strong performance in medical-image and remote-sensing understanding. Overall, an increasing range of previously specialized vision tasks is becoming accessible through general-purpose models.

\textbf{Task Properties Associated With the Remaining Gap.}\, The remaining specialist gaps are associated with requirements for precise geometry, temporal consistency, faithful reconstruction, and fine-grained domain knowledge. In 3D perception, the larger gap appears when spatial understanding must become quantitatively accurate and geometrically consistent: depth estimation (See \Cref{fig:astra_3d_tasks}) and multiview reconstruction remain sensitive to metric scale, local surface geometry, camera motion, and alignment across views. In video segmentation, the remaining difficulty is concentrated in precise boundaries, small structures, separation of nearby instances (See \Cref{fig:davis_qualitative}), and maintaining these distinctions consistently across frames, indicating that dense temporal prediction remains less mature than higher-level video understanding. Image restoration exposes a different limitation, where visually plausible improvement does not necessarily correspond to faithful recovery of the original image; fine textures and edges may be altered or re-synthesized, while some degradations remain insufficiently corrected. Expert-domain tasks introduce an additional requirement for specialized visual knowledge (See \Cref{fig:medical}). In pathology, nuclei can be localized accurately, but assigning the correct nucleus type remains substantially harder, particularly for subtle or less frequent categories.

\begin{figure}[t]
  \centering
  \includegraphics[width=\linewidth]{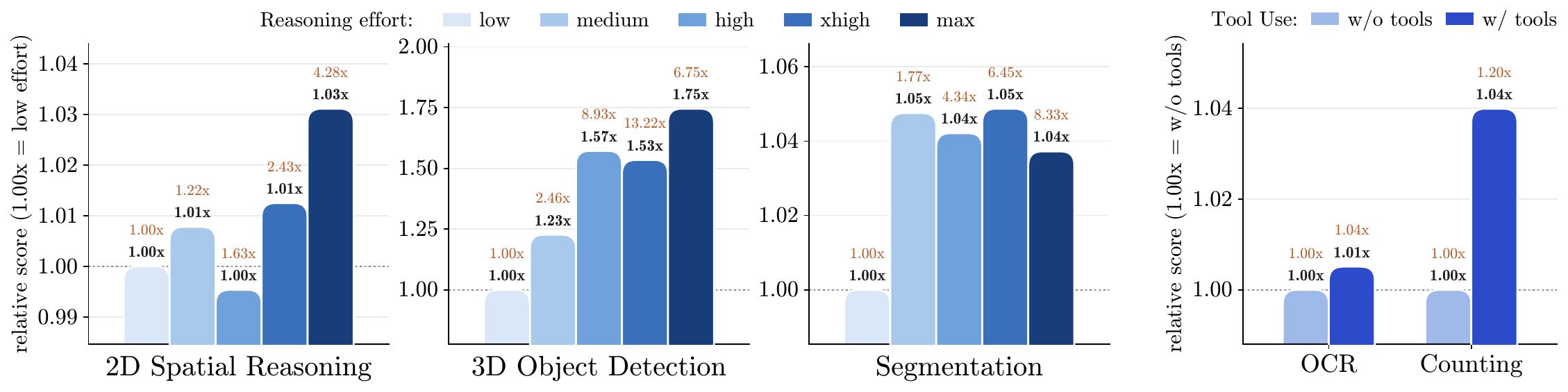}
  \caption{\textbf{Effect of reasoning effort and tool use on GPT-6 Astra.} The left three plots compare reasoning-effort levels for 2D spatial reasoning, 3D object detection, and segmentation. The annotations above each bar report the \textbf{score ratio} (black) and \textbf{inference-cost ratio} (red), both computed relative to the \textit{low-effort} baseline. The right panel compares runs with and without tools at \texttt{xhigh} effort for OCR and counting, using the run without tools as the baseline for both ratios.}
  \label{fig:effort_tools}
\end{figure}

\section{Can Reasoning and Tools Close the Gap?}
\researchquestion[More reasoning and tool use can close selected specialist gaps, but gains are task-dependent and often come with substantially higher inference cost.]{Question 4}{\textbf{Can added reasoning, explicit tool use, or open specialist-as-tool pipelines close these gaps, and at what cost?}
}
We observe that additional reasoning and tool use can close selected gaps with specialist models (See~\Cref{fig:effort_tools}). Higher reasoning brings 2D spatial reasoning, 3D object detection, and image generation and editing to specialist-level performance. In 3D object detection, for example, a system may recognize an object correctly but still need more reasoning to estimate its position, size, and orientation. Tools also improve chart and document understanding, counting, and 2D medical grounding; in counting, for example, the system first points to individual objects and then counts the identified instances. Segmentation also benefits from additional reasoning, although the gains plateau at higher effort and remain insufficient to close the specialist gap. A pattern across these comparisons is that additional effort produces smaller gains in some already strong perception and visual reasoning settings, while more useful gains appear where the system shows an emerging capability but applies it inconsistently. 

The cost of these improvements varies substantially. Tool use can yield gains at modest additional cost (1.04–1.20×), while higher reasoning can require several times the baseline cost (up to 13.22×), sometimes for only small improvements. Specialist models offer a complementary route when larger gaps remain, supplying visual predictions that the generalist can use to complete the task. For example, segmentation models identify individual objects for counting, while depth models help the system compare distances for spatial reasoning. Dedicated grounding models can substantially improve localization in remote-sensing imagery, where fine-scale targets remain difficult for the generalist to identify accurately. Image generation and editing follow a similar division of work: the reasoning model interprets the request and directs an image model to produce the required output. In these settings, progress comes from combining the generalist’s understanding of the task with the specialist’s ability to provide the visual information or output needed to carry it out.

\section{Where Does General-Purpose Vision Stand?}
\begin{figure}[t]
  \centering
  \includegraphics[width=0.8\linewidth]{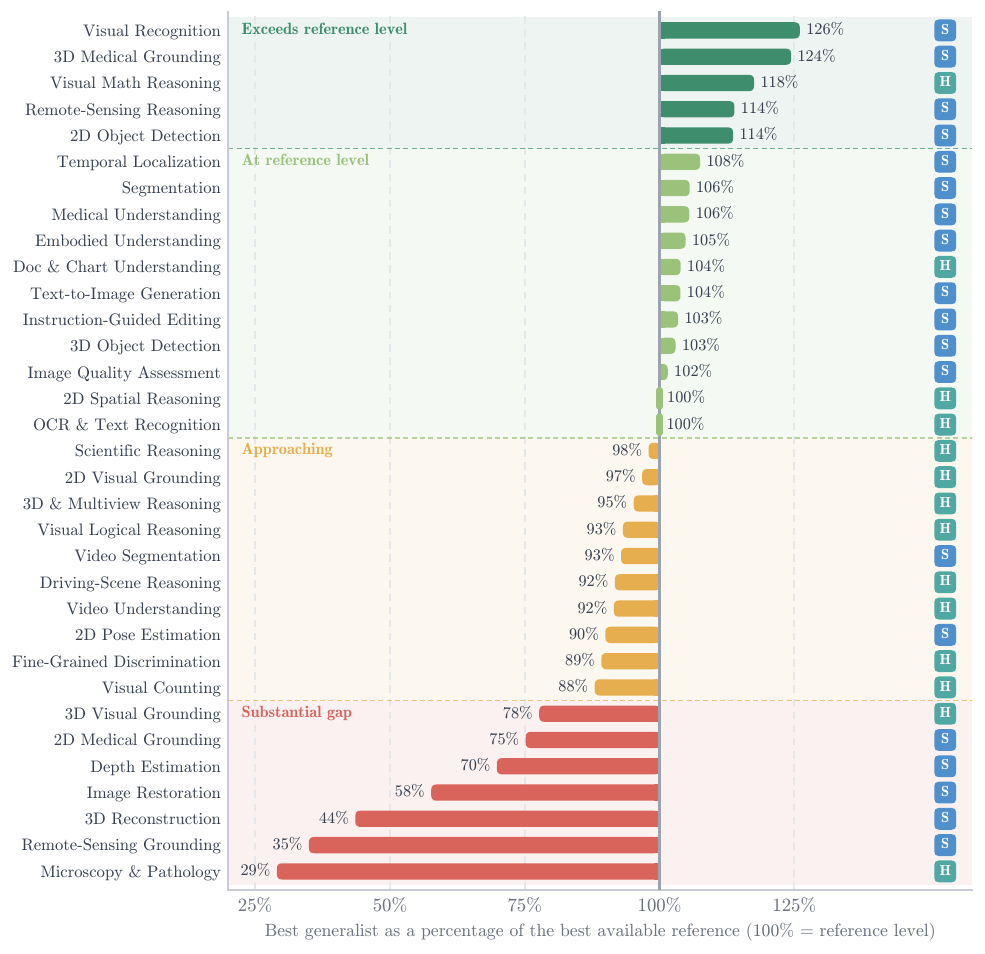}
  \caption{\textbf{Capability maturity relative to reference levels.} For each capability with a reference, the best generalist score as a percentage of the stronger of the specialist (S) and human (H) references, with the ratio inverted for lower-is-better metrics. Capabilities are grouped into four tiers: exceeds reference level (above 110\%), at reference level (100--110\%), approaching (85--100\%), and substantial gap (below 85\%).}
  \label{fig:tier_bands}
\end{figure}

To understand how far frontier visual capabilities have progressed, we consider how close their performance is to established reference levels. We group capabilities into \textbf{four maturity tiers}: \textit{exceeds reference level}, \textit{at reference level}, \textit{approaching}, and \textit{substantial gap}, using the best generalist performance for each capability. 
Depending on the task, the reference comes from a dedicated specialist, a prior generalist SOTA, or human performance. Human performance is particularly useful where a suitable model reference is unavailable or where the best evaluated generalist surpasses the selected model reference and a stronger comparison is needed to assess the remaining headroom. \Cref{fig:tier_bands} brings these comparisons together to provide an overview of capability maturity across the evaluated tasks.

For several capabilities, spanning visual recognition, visual reading, reasoning, localization, and image generation, the best evaluated generalist reaches or exceeds the selected reference. These results show that general-purpose models can increasingly support tasks traditionally handled by dedicated models. However, progress remains uneven. Some capabilities are approaching reference performance, while others still show substantial gaps, particularly where precise geometry, faithful reconstruction, or specialized visual knowledge is required. The choice of reference also affects how we interpret these results. For example, generalists surpass the evaluated specialist in 3D visual grounding but remain substantially below human performance, leaving considerable room for improvement. Together, these results show which capabilities are becoming accessible through general-purpose models and where further advances are needed. They also raise a question for future systems: \textit{which visual capabilities should be internalized by a generalist, and which are better provided through tools?} Our results suggest a tentative division of work. Semantic interpretation, language-conditioned reasoning, and task planning and coordination are natural capabilities to strengthen within the generalist. Specialist tools may remain particularly valuable for metric geometry, faithful reconstruction, dense correspondence, high-fidelity restoration, and fine-grained discrimination in rare expert domains. A key challenge is to determine when the generalist should act directly, when it should call a specialist, and how it should verify the resulting output, while accounting for accuracy, cost, and latency. As general-purpose models become more capable, such an assessment can help the field understand what has improved, what remains difficult, and where future research can make the greatest difference.

\section{Additional Evaluation Details}
\label{app:evaluation-details}

\begin{table}[!h]
\centering
\small
\begin{tabular}{>{\raggedright\arraybackslash}p{0.28\linewidth}>{\raggedright\arraybackslash}p{0.17\linewidth}>{\raggedright\arraybackslash}p{0.47\linewidth}}
\toprule
Capability & Metric & Specialist Model \\
\midrule
Visual Recognition & Accuracy & SEAL~\citep{wu2024v}; Gemini 3 Pro~\citep{google2025gemini3provision} \\
\midrule
2D Object Detection & F1@IoU=0.5 & Rex-Omni~\citep{jiang2026detect} \\
Segmentation & gIoU & SAM 3 Agent~\citep{ICLR2026_e0982cbc} \\
2D Pose Estimation & OKS AP & ViTPose++-L~\citep{10308645} \\
\midrule
Depth Estimation & AbsRel $\downarrow$ & DA3~\citep{lin2025depth3recoveringvisual} \\
3D Object Detection & AP3D & WildDet3D~\citep{huang2026wilddet3dscalingpromptable3d} \\
3D Visual Grounding & Acc@IoU0.25 & Gemini-2.5-Pro~\citep{wang2026objects} \\
3D Reconstruction & Error $\downarrow$ & VGGT-1B~\citep{wang2025vggt} \\
\midrule
Video Understanding & Accuracy & GPT4-o1~\citep{jaech2024openai,zhao2025mmvu}; Cambrian-S-7B~\citep{ICLR2026_7e3dcf77} \\
Temporal Localization & R@1 IoU=0.7 & TRACE~\citep{guo2025tracetemporalgroundingvideo} \\
Video Segmentation & J\&F & BEE~\citep{HOU2026114333} \\
\midrule
Text-to-image Generation & Soft-TIFA GM & Structured Conditioning + Qwen-Image~\citep{wu2025qwenimagetechnicalreport} \\
Instruction-guided Editing & Score (1--5) & Boogu-Image-0.1-Edit-Thinking~\citep{chen2026booguimage01boostingopenagentic} \\
Image Restoration & PSNR & MIRAGE~\citep{ICLR2026_cd1da804} \\
Image Quality Assessment & Accuracy & CoInstruct~\citep{10.1007/978-3-031-72646-0_21}; UniPercept~\citep{cao2025uniperceptunifiedperceptuallevelimage} \\
\midrule
Driving-scene Reasoning & Accuracy & Qwen-Drive-1.0-SFT~\citep{zhou2026qwendrive10initialstepvisionlanguage}; GPT4-o1~\citep{jaech2024openai,corbière2025retrievalbasedinterleavedvisualchainofthought} \\
Embodied Understanding & Accuracy & Gemini Robotics-ER 2~\citep{deepmind2026geminiroboticser2} \\
\midrule
Medical Understanding & Accuracy & GPT-5.6 Sol~\citep{openai2026gpt56blog,qwen38} \\
2D Medical Grounding & mAP@0.5 & RadVLM~\citep{deperrois2026radvlm} \\
3D Medical Grounding & mIoU & M3D-LaMed-Llama-2-7B~\citep{bai2024m3dadvancing3dmedical} \\
Microscopy \& Pathology & Macro-F1 & HoVer-NeXt~\citep{TORBATI2026100933} \\
Remote-sensing Reasoning & Accuracy & GPT-5.4~\citep{openai2026gpt54,luo2026vlrsbenchvisionlanguagereasoningbenchmark} \\
Remote-sensing Grounding & mIoU & RSRefSeg 2~\citep{11313649} \\
\bottomrule
\end{tabular}
\caption{Specialist reference model for each capability.}
\label{tab:specialist_models}
\end{table}

\textbf{Benchmarks}. \, We prioritize challenging benchmarks that retain meaningful headroom for current frontier systems while collectively covering a broad range of computer-vision tasks. We organize their content into capabilities, drawing on complete benchmarks, relevant subsets of their tasks, or combinations of benchmarks as appropriate. Our evaluation includes 
MMStar~\citep{chen2024we}, BabyVision~\citep{chen2026babyvision}, BLINK~\citep{fu2024blink}, V*Bench~\citep{wu2024v}, PerceptionBench~\citep{lin2026perceptionbench}, WorldVQA~\citep{zhou2026worldvqa}, RealWorldQA~\citep{xai2024realworldqa}, BlindTest~\citep{rahmanzadehgervi2024vision}, MMMU-Pro~\citep{yue2025mmmu}, VisualPuzzles~\citep{song2025visualpuzzles}, ZeroBench~\citep{roberts2025zerobench}, MathVista~\citep{lu2024mathvista}, MathVision~\citep{wang2024measuring}, PixMo-Count~\citep{deitke2025molmo}, CountQA~\citep{tamarapalli2025countqa}, VLMsAreBiased~\citep{vo2025vision}, 
InfoVQA~\citep{mathew2022infographicvqa}, CharXiv~\citep{wang2024charxiv},
MindCube-Tiny~\citep{wang2025mindcube}, OmniSpatial~\citep{jia2026omnispatial}, VisFactor~\citep{huang2025visfactor}, MMSI-Bench~\citep{yang2026mmsi},
Dense200~\citep{jiang2026detect}, ScreenSpotPro~\citep{li2025screenspot}, RefCOCO~\citep{kazemzadeh2014referitgame}, ReasonSeg~\citep{lai2024lisa}, OCHuman~\citep{8953934},
Anywhere3Dv2~\citep{wang2026objects}, DIODE~\citep{vasiljevic2019diode}, ETH3D~\citep{schoeps2017cvpr}, Omni3D~\citep{brazil2023omni3d}, 
MMVU~\citep{zhao2025mmvu}, VSI-Bench~\citep{yang2025thinking}, ActivityNet~\citep{krishna2017dense_anet}, DAVIS~\citep{pont20172017},
ERQA~\citep{team2025gemini_erqa}, LingoQA~\citep{10.1007/978-3-031-72980-5_15}, DrivingVQA~\citep{corbière2025retrievalbasedinterleavedvisualchainofthought}, EmbodiedBench~\citep{yang2025embodiedbenchcomprehensivebenchmarkingmultimodal},
MedXpertQA-MM~\citep{zuo2025medxpertqabenchmarkingexpertlevelmedical}, MMMU-Pro-Med~\citep{yue2025mmmu}, MS-CXR~\citep{10.1007/978-3-031-20059-5_1,PhysioNet-ms-cxr-1.1.0,pollard_physionet_2026}, M3D-Bench~\citep{bai2024m3dadvancing3dmedical}, PUMA~\citep{schuiveling2025novel}, 
VLRS-Bench~\citep{luo2026vlrsbenchvisionlanguagereasoningbenchmark}, RefSegRS~\citep{yuan2024rrsisreferringremotesensing},
Q-Bench2~\citep{zhang2024q}, UniPercept~\citep{cao2025unipercept}, BSD68~\citep{937655}, Urban100~\citep{Huang_2015_CVPR}, Rain100L~\citep{9157061}, SOTS~\citep{8451944}, GoPro~\citep{Nah_2017_CVPR}, LOL-v1~\citep{wei2018deepretinexdecompositionlowlight}, GenEval2~\citep{kamath2025geneval} and ImageEditBench~\citep{ye2026imgedit}.

\textbf{Output processing}.\, We convert model predictions into the output formats required by each benchmark. Segmentation polygons are rasterized into pixel masks, while some tool-based runs produce masks directly. For depth estimation, predicted surfaces and depth anchors are rendered into dense depth maps, or the maps are generated directly by executing model-written code. For 3D reconstruction, depth predictions are combined with camera parameters to obtain point clouds or per-frame point maps in a shared coordinate system. For 3D detection and grounding, predicted centers, dimensions, and rotations, where applicable, are converted into box boundaries or cuboid corners. For microscopy and pathology, instance and class maps are converted into labeled nucleus polygons; nucleus centers are used for detection scoring and tissue masks for segmentation scoring. For DPG-Bench, four generated images are assembled into the grid expected by the evaluator.

\textbf{Image generation and editing}.\, We pair GPT-6 Astra with GPT Image 2.5 Sunburst~\citep{openai2026gptimagesunburst}, Qwen 3.8-Max with Qwen Image 3 Pro~\citep{qwen2026qwenimage3}, Muse Spark 1.3 with Muse Image~\citep{meta2026museimage}, and Gemini 3.1 Pro with Gemini 3.1 Flash Image~\citep{google2026gemini31flashimage}. In each case, the reasoning model formulates the instructions, while the corresponding image model generates or edits the image.
\textbf{Reasoning configurations}.\, We test multiple reasoning-effort settings, including high, xhigh, and max, wherever supported by the corresponding system and interface. These comparisons examine how additional reasoning affects task performance and inference cost.
\textbf{Specialist models.}\, The specialist scores reported in the main results ~\Cref{tab:capabilities-abs} correspond to different models across capabilities. We provide a breakdown of the specialist models in~\Cref{tab:specialist_models} and the capabilities for which they are used.

\section{Conclusion}
We evaluated the recent GPT-6 Astra alongside five frontier general-purpose systems across 34 capabilities and 55 benchmarks. 
The results show how far language-model-driven general-purpose systems have expanded across the computer vision landscape.
Astra demonstrates strong capabilities across visual reasoning, structured prediction, 3D perception, video, and expert domains, with several tasks approaching or reaching available reference levels. 
Yet these advances are not uniform: some of Astra’s strongest capabilities remain substantially less developed in other frontier systems.
More broadly, our results suggest that \textit{what is hard in computer vision} is changing. 
General-purpose systems increasingly succeed when visual information can be interpreted, reasoned over, or organized around objects.
Larger gaps remain when tasks demand precise metric geometry, faithful reconstruction, temporal consistency, or specialized fine-grained visual knowledge.
Rather than simply replacing specialized vision, general-purpose models are redrawing the boundary between generalist and specialist capabilities, a boundary that will continue to evolve as these systems become more capable.
\bibliographystyle{assets/plainnat}
\bibliography{resources/main}

\end{document}